\documentclass[fleqn,10pt]{wlscirep}
\usepackage[utf8]{inputenc}
\usepackage[T1]{fontenc}
\usepackage{lineno}

\usepackage{hyperref}       
\usepackage{url}            
\usepackage{booktabs}       
\usepackage{amsfonts}       
\usepackage{nicefrac}       
\usepackage{microtype}      
\usepackage{xcolor}         
\usepackage{multicol}         
\usepackage{multirow}         
\usepackage{adjustbox}
\usepackage{floatrow}
\usepackage{algorithmic}
\usepackage[linesnumbered, ruled, vlined]{algorithm2e}

\title{A Multi-scale Time-series Dataset with Benchmark for Machine Learning in Decarbonized Energy Grids}

\author[1,$\dag$]{Xiangtian Zheng}
\author[2,$\dag$]{Nan Xu}
\author[2,$\dag$]{Loc Trinh}
\author[1,$\dag$]{Dongqi Wu}
\author[1]{Tong Huang}
\author[1]{S. Sivaranjani}
\author[2,*]{Yan Liu}
\author[1,3,*]{Le Xie}
\affil[1]{Texas A\&M University, Department of Electrical and Computer Engineering, College Station, 77840, USA}
\affil[2]{University of Southern California, Computer Science Department, Los Angeles, 90007, USA}
\affil[3]{Texas A\&M Energy Institute, College Station, 77840, USA}
\affil[*]{corresponding author(s): Le Xie (le.xie@tamu.edu) and Yan Liu (yanliu.cs@usc.edu)}

\affil[$\dag$]{these authors contributed equally to this work}

\begin{abstract}
The electric grid is a key enabling infrastructure for the ambitious transition towards carbon neutrality as we grapple with climate change. With deepening penetration of renewable resources, the reliable operation of the electric grid becomes increasingly challenging. In this paper, we present PSML, a first-of-its-kind open-access multi-scale time-series dataset, to aid in the development of data-driven machine learning (ML)-based approaches towards reliable operation of future electric grids. The dataset is synthesized from a joint transmission and distribution electric grid to capture the increasingly important interactions and uncertainties of the grid dynamics, containing power, voltage and current measurements over multiple spatio-temporal scales. Using PSML, we provide state-of-the-art ML benchmarks on three challenging use cases of critical importance to achieve: (i) early detection, accurate classification and localization of dynamic disturbances; (ii) robust hierarchical forecasting of load and renewable energy; and (iii) realistic synthetic generation of physical-law-constrained measurements. We envision that this dataset will provide use-inspired ML research in safety-critical systems, while simultaneously enabling ML researchers to contribute towards decarbonization of energy sectors. 
\end{abstract}
\begin{document}

\flushbottom
\maketitle

\thispagestyle{empty}


\section*{Background \& Summary}
The electric grid is one of the largest sources of carbon emissions, and is expected to play a key role in tackling climate change~\cite{climate_and_change}. The electricity sector around the world is undergoing a major transition towards carbon neutrality with deepening penetration of renewable energy resources and vehicle electrification. The variability of renewable energy resources along with growing electricity demand and system vulnerability under extreme weather events pose pressing technological challenges during this transition~\cite{xie2021toward}. Conventional physics-based modeling, optimization and control tools are becoming inadequate in these evolving systems due to the high degree of uncertainty and variability in power generation,  consumption, and environmental factors such as climate change. 

During this period of energy sector transition, there are enormous opportunities for artificial intelligence (AI) and machine learning (ML)-based methods~\cite{rolnick2019tackling} to improve grid operations ranging from more accurate forecasting of renewables and load~\cite{8464297,6945846,9068279}, to planning~\cite{9069289,9479716,9454298}, real-time monitoring \cite{6808416,9043670,huang2021neural}, control~\cite{9420757,8873679} and protection~\cite{9029268}. Conversely, power systems are highly nonlinear dynamical systems with interesting physical phenomena over various time scales; indeed, we believe that the breadth of problems available in this domain can stimulate the development of new algorithms, tools, and techniques in ML. 

In order to foster advances that are mutually beneficial to both the ML and power system communities, it is necessary to develop well-documented and calibrated open-source datasets and use cases that are relevant to real-world power engineering problems, while simultaneously being accessible and usable to ML researchers with limited backgrounds in power/energy systems. There have been attempts at developing ML benchmarks for various power system tasks such as renewable~\cite{voyant2017machine,dolara2018comparison} and load forecasting~\cite{yildiz2017review,feng2020assessment,almalaq2017review}, and fault and anomaly detection~\cite{zainab2019faulted,cui2019machine,hink2014machine,mohammadpourfard2019benchmark}. Other researchers have attempted to accelerate algorithm development by providing online simulation platforms for specific tasks, such as the L2RPN competition~\cite{marot2020learning,marot2021learning} and the oscillation source location contest~\cite{OLS}. 

However, the development of a cross time-and-spatial scale, open-source dataset from the power engineering domain that can be utilized by the broader ML community is still at a nascent stage, with several gaps in existing sources. Firstly, most benchmarks for ML in power systems employ datasets that are either scattered across multiple independent system operators, as in the case of load and renewable data, or not publicly available, as in the case of dynamic data. Secondly, the identification of relevant problems, dataset generation, and implementation of dedicated ML-based algorithms, all require deep knowledge of the power engineering domain and diverse power system simulation tools. This lack of coherent comprehensive datasets along with well-defined tasks is the key barrier for ML communities to contribute to power system problems. Finally, there is a lack of consistent domain-relevant assessment metrics against which different ML algorithms can be compared. 


In this paper, we bridge this gap by creating a comprehensive open-source dataset along with associated use cases and benchmarks that are relevant to the power system research community. The dataset contains minute-level real-world load, weather and renewable time series data over 3 years from 66 areas across the U.S., minute-level synchrophasor measurements of 1 year in 3 scenarios, and millisecond-level synchrophasor measurements in more than 1000 disturbance cases. This is synthesized from a joint transmission (bulk) and distribution (retail) electric grid that contains a rich and diverse set of energy resources and dynamic events. This dataset is self-contained and coherent across transmission and distribution-level dynamics at multiple time-scales, making it ideally suited for developing multi-scale ML-based algorithms. We then benchmark the performance of both traditional algorithms from the power engineering domain and common ML-based algorithms for three tasks of critical importance to the power engineering community, namely forecasting, monitoring, and simulating. Specifically, we consider the following use cases: (i) early detection, accurate classification and localization of dynamic disturbance events; (ii) robust hierarchical forecasting of load and renewable energy; and (iii) realistic synthetic generation of physical-law-constrained measurement time series.



In summary, we develop one of the first comprehensive open-source datasets with associated use cases and benchmarks from the power systems domain that can be leveraged by ML researchers interested in advancing the state-of-the-art in time series forecasting, classification, and generation, while contributing towards future zero-carbon energy systems. The full dataset and benchmark codes can be downloaded from GitHub~\cite{PSML_github_code} and Zenodo~\cite{PSML_dataset}. 

\section*{Methods}
\begin{figure}[hbpt!]
    \centering
    \includegraphics[width=1.0\textwidth]{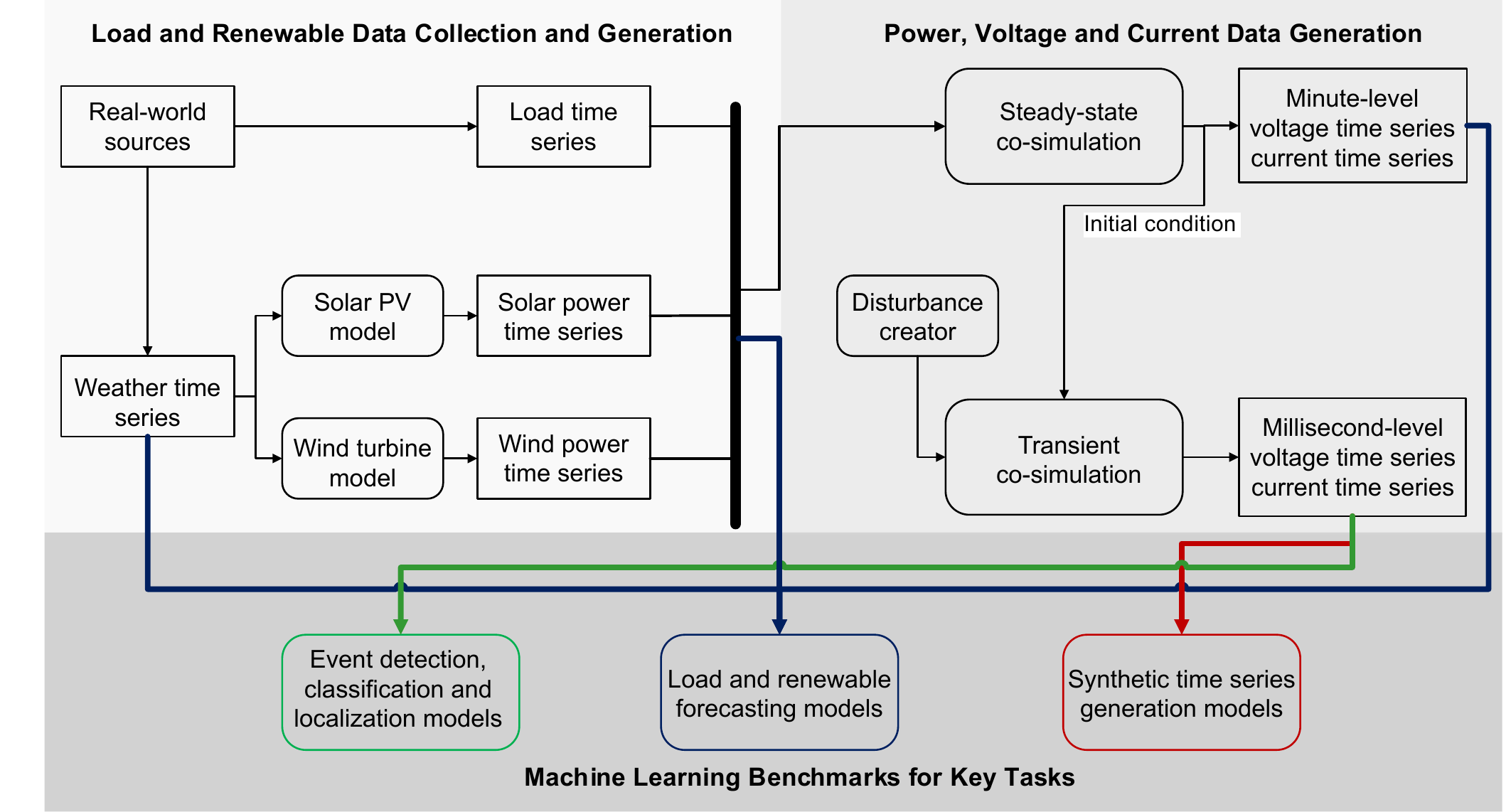}
    \caption{Conceptual block diagram of the data flow of the transmission + distribution co-simulation platform used to create \texttt{PSML}. While the simulation is a closely integrated process that combines all types of input data, results with different time-scales are generated at different simulation stages.}\label{fig:cosimulation-flowchart}
\end{figure}

In this section, we describe the method used to create the multi-scale time-series dataset along with the benchmark models, which includes co-simulation model development, data generation methods, and benchmark algorithms for key tasks.
A brief workflow overview is shown in Fig. \ref{fig:cosimulation-flowchart}, including (i) load and renewable data collection and generation, (ii) power, voltage and current data generation, and (iii) machine learning benchmarks for key tasks. The process starts with collecting the real-world weather and load time series data. We then generate the solar and wind generation profiles based on the physical renewable generation models given the corresponding weather data. To obtain the multi-scale measurement data, we conduct steady-state power flow simulation under different system conditions of load and renewable generation and perform transient dynamic simulation under various random disturbances by a novel joint transmission and distribution (T+D) grid simulation platform. We then benchmark the performance of traditional algorithms from the power engineering domain and common ML-based algorithms for selected key tasks of interests.
Details about the methods for each step are elaborated in the rest of this section. Details about the source code and method implementations are explained in the Code availability section.

\textbf{Remark:}
The voltage and current measurement time series are simulation data. The reason that we include simulation data in the PSML dataset consists of three aspects. First, real operational data of the power grid are typically confidential and most are forbidden to be publicly shared, due to policies such as critical energy/electric infrastructure information (CEII). Second, some high-impact events that are challenging for analysis, such as forced oscillations under resonance conditions, are rarely observed in real-world operational data. Considering such a small amount of challenging events is insufficient for the training, testing, and validation of ML algorithms. Third, while real measurement data can reflect the impact of a small amount of renewable energy in today's power systems, it cannot capture the dynamics of the future grid with deep renewable penetration.

\subsection*{Co-simulation Model Development}
\begin{figure}[hbpt!]
    \centering
    \includegraphics[width=0.9\textwidth]{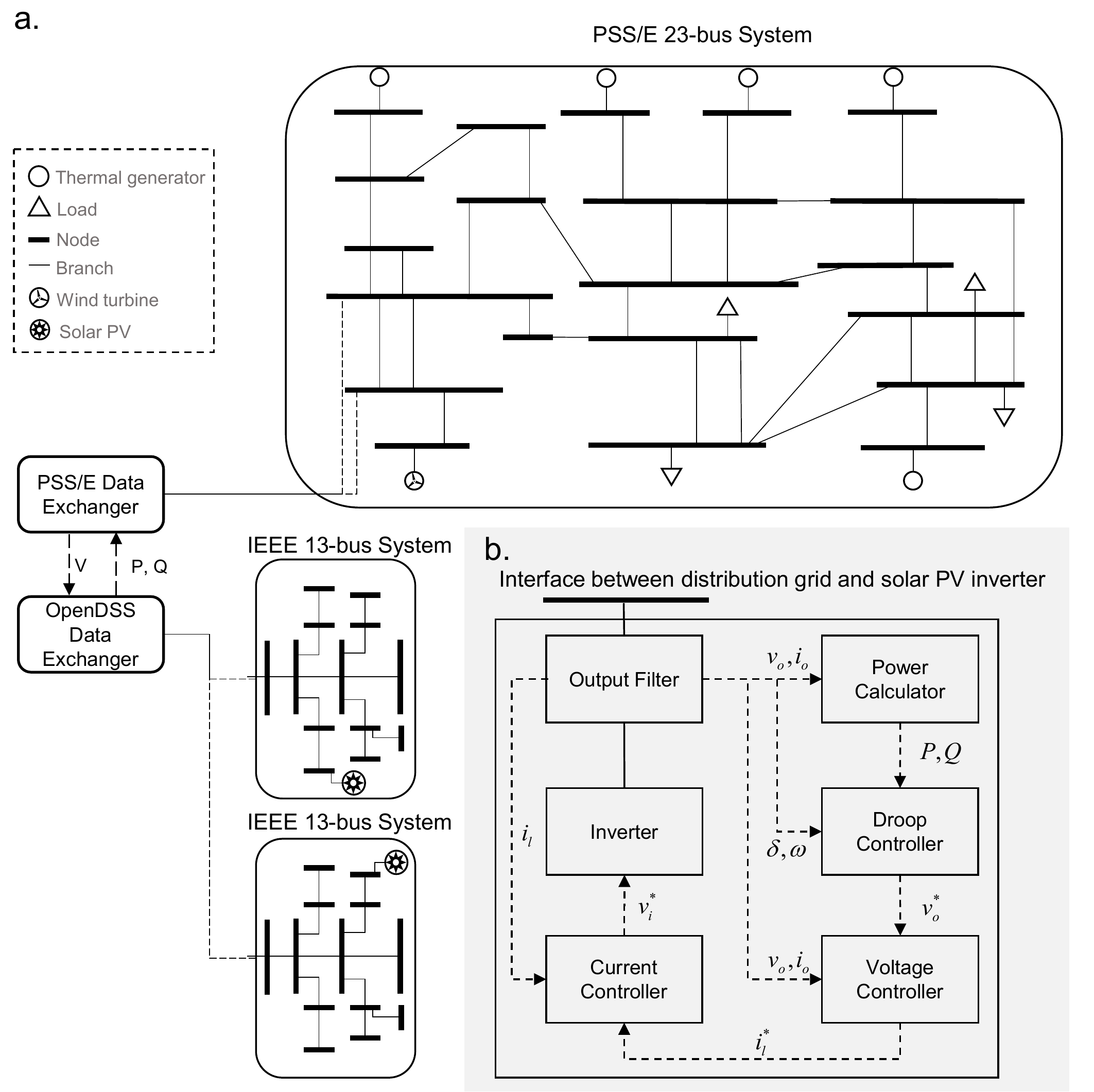}
    \caption{Diagram of the join transmission and distribution (T+D) simulation model. a. One PSS/E 23-bus system and two IEEE 13-bus systems implement the co-simulation by exchanging the real-time voltage and power information, which are respectively simulated by the PSS/E and OpenDSS. b. Details of the inverter-based solar PV dynamic model in the distribution system.}
    \label{fig:appendix:cosimulation}
\end{figure}

We create a joint T+D grid simulation platform that consists of one PSS/E 23-bus transmission system~\cite{psse} and two IEEE 13-bus distribution systems~\cite{IEEEfeeder} as shown in Fig. \ref{fig:appendix:cosimulation}-a.
The model of the bulk transmission grid is implemented by PSS/E, modified from the original PSS/E 23 bus test system that has 6 thermal generators and 7 load buses. We replace one thermal generator with a wind turbine model to represent renewable-rich scenarios. We also connect two load buses to the distribution grid distribution systems while keeping the rest of the load buses connected to lumped load. To better comply with the ratings and size of the benchmark distribution systems, all load capacities are reduced by roughly 40\% of their original values. The model of the distribution grids is the IEEE 13-bus feeder \cite{IEEEfeeder} implemented by OpenDSS, which are connected to the corresponding load buses in the transmission system model. In each distribution grid, we respectively attach solar photovoltaic (PV) and power inverter models to load buses, which represents aggregated residential roof-top solar generation.

As OpenDSS has no template of solar generation dynamic models, we build an inverter-interfaced solar PV dynamic model in Python scripts that can be used as a customized dynamic model compatible with OpenDSS. Following the literature~\cite{huang2021neural2}, the model consists of several typical controller components as shown in Fig. \ref{fig:appendix:cosimulation}-b and the dynamics are summarized as follows. The dynamics of the power calculator follows Eq.~\ref{power_calculator_1} and~\ref{power_calculator_2}, where $i_\text{o}$ and $v_\text{o}$ are instantaneous current and voltage at the terminal of the output filter, $P$ and $Q$ are the active and reactive output power, $\omega_\text{c}$ is the upper frequency of the low-pass output filter, and $i_{\text{od}}$, $i_{\text{oq}}$, $v_{\text{od}}$, and $v_{\text{oq}}$ are respectively the direct and quadrature components of $i_o$ and $v_o$.
The dynamics of the frequency droop controller follows Eq.~\ref{eq:droop_controller_1}-\ref{eq:droop_controller_4}, where $\omega_0$ is the nominal frequency, $P^*$, $Q^*$ and $E^*$ are the dispatched set points of the active power, reactive power and potential, $v_\text{d}^*$ and $v_\text{q}^*$ is the set point of the subsequent voltage controller, and $M_\text{f}$, $D_\text{f}$, $M_\text{v}$ and $D_\text{v}$ are the predefined parameters of the controller. Specifically,  Eq.~\ref{eq:droop_controller_1} and~\ref{eq:droop_controller_2} describe the dynamics mimicing traditional synchronous generators while Eq.~\ref{eq:droop_controller_4} describe the way how the droop controller adapts the set point of the subsequent voltage controller.
The dynamics of the voltage and current controller is elaborated by Eq.~\ref{eq:vc_controller_1}-\ref{eq:vc_controller_8}, where $\xi_\text{d}$ and $\xi_\text{q}$ are the state variables of the current controller, $i_{\text{ld}}^*$ and $i_{\text{lq}}^*$ are the set points of the current controller, $v_{\text{id}}^*$ and $v_{\text{iq}}^*$ are the set points of the inverter, $K_{\text{iv}}$, $F$ and $K_{\text{pv}}$ are the predefined parameters of the voltage controller, $K_{\text{ic}}$ and $K_{\text{pc}}$ are the predefined parameters of the current controller, and $C_\text{f}$ and $L_\text{f}$ are the capacitance and inductance of the output filter.
The dynamics of the output filter can be described by Eq.~\ref{eq:filter_1}-\ref{eq:filter_4}, where $r_\text{f}$, $C_\text{f}$ and $L_\text{f}$ are the resistance, capacitance and inductance of the output filter.
\begin{align}
        &\dot{P}=-\omega_\text{c} P+\omega_\text{c}(v_{\text{od}}i_{\text{od}}+v_{\text{oq}}i_{\text{oq}})\label{power_calculator_1}\\
    &\dot{Q}=-\omega_\text{c} P+\omega_\text{c}(v_{\text{od}}i_{\text{od}}+v_{\text{oq}}i_{\text{oq}})\label{power_calculator_2}\\
        & \dot{\delta}=\omega-\omega_0\label{eq:droop_controller_1}\\
    & M_\text{f}\dot{\omega}=-D_\text{f}(\omega-\omega_0)+P^*-P\label{eq:droop_controller_2}\\
    & M_\text{v}\dot{v}_{\text{od}}^*=D_\text{v}(Q^*-Q)-(v_{\text{od}}^*-E^*),\quad v_{\text{oq}}^*=0\label{eq:droop_controller_4}\\
        &\dot{\xi}_\text{d}=v_{\text{od}}^*-v_{\text{od}},\quad \dot{\xi}_\text{q}=v_{\text{oq}}^*-v_{\text{oq}}\label{eq:vc_controller_1}\\
    &\dot{\psi}_\text{d}=i_{\text{ld}}^*-i_{\text{ld}},\quad \dot{\psi}_\text{q}=i_{\text{lq}}^*-i_{\text{lq}}\label{eq:vc_controller_3}\\
    &i_{\text{ld}}^*=K_{\text{iv}}\xi_\text{d}+Fi_{\text{od}}+K_{\text{pv}}(v_{\text{od}}^*-v_{\text{od}})-\omega_0C_\text{f}fv_{\text{oq}}\label{eq:vc_controller_5}\\
    &i_{\text{lq}}^*=K_{\text{iv}}\xi_\text{q}+Fi_{\text{oq}}+K_{\text{pv}}(v_{\text{oq}}^*-v_{\text{oq}})-\omega_0C_\text{f}fv_{\text{od}}\label{eq:vc_controller_6}\\
    &v_{\text{id}}^*=K_{\text{ic}}\psi_{\text{d}}+K_{\text{pc}}(i_{\text{ld}}^*-i_{\text{ld}})-\omega_0L_{\text{f}}i_{\text{lq}} \label{eq:vc_controller_7}\\
    &v_{\text{iq}}^*=K_{\text{ic}}\psi_{\text{q}}+K_{\text{pc}}(i_{\text{lq}}^*-i_{\text{lq}})-\omega_0L_{\text{f}}i_{\text{ld}} \label{eq:vc_controller_8}\\
        &L_\text{f}\dot{i}_\text{ld}=-r_\text{f}i_\text{ld}+\omega L_\text{f}i_\text{lq}+v_\text{id}-v_\text{od}\label{eq:filter_1}\\
    &L_\text{f}\dot{i}_\text{lq}=-r_\text{f}i_\text{lq}+\omega L_\text{f}i_\text{ld}+v_\text{iq}-v_\text{oq}\label{eq:filter_2}\\
    &C_\text{f}\dot{v}_\text{od}=\omega C_\text{f}v_\text{oq}+i_\text{ld}-i_\text{od}\label{eq:filter_3}\\
    &C_\text{f}\dot{v}_\text{oq}=\omega C_\text{f}v_\text{od}+i_\text{lq}-i_\text{oq}\label{eq:filter_4}
\end{align}

To implement the T+D co-simulation, we further create a Python control process to facilitate workflow coordination and data exchange between PSS/E, OpenDSS, PV inverter models and input data files. Specifically, the python control process is able to (i) start and pause the simulation in the PSS/E and OpenDSS every step, (ii) read and store the updated variables of the transmission and distribution grid models after each step, and (iii) overwrite the values of certain variables at the beginning of each step. In such way, we can exchange the voltage and power data between the transmission and distribution systems step by step that enables the T+D co-simulation. Please refer to the Code availability section for the details about the source code and method implementations.

\subsection*{Data Generation}
This subsection introduces the methods how we collect the real-world data and generate the measurement data via co-simulation. The source code of the data collection and simulation implementation can be found in the Github repository as elaborated in the Code availability section.
\subsubsection*{Load and Renewable Time Series Data Generation}
We collect real-world load time series and synthesize active power time series of renewable generation along with real-world weather data.
For collecting the load power time series, we aggregate hourly real-world load data of  representative 66 load zones ranging from 2018 to 2020, obtained from major power markets in the U.S. that regulate about 70\% of U.S. electricity sales \cite{ruan2020cross}.
To incorporate renewable power and weather time series in \texttt{PSML}, we collect real weather data of 5-minute resolution of each load zone from 2018 to 2020 from National Solar Radiation Database (NSRDB) \cite{NSRDB}. The selected weather station of each load zone locates around major cities within range. We calculate the renewable generation power based on the collected weather data of each load zone. The active power output of residential solar photovoltaic (PV) is estimated by the System Advisor Model (SAM) \cite{blair2014system}, based on the solar radiation-related data. The active power output of wind turbines is estimated by the location-dependent wind turbine power curves \cite{EIA} based on the collected wind-related data.
Finally, we aggregate time-stamped load, renewable and weather data of 66 load zones by interpolation in the \texttt{PSML} dataset. For the convenience of subsequent simulation, the load data are further normalized by their 3-year average value, while the renewable data are normalized by the nominal power values of the physical renewable models.

\subsubsection*{Minute-level Voltage, Current and Power Time Series} 
The steady-state simulation produces minute-level data of the power transmission and distribution system using the load and renewable profiles. These profiles specifies the net real and reactive power consumption across the system except the buses where dispatch-able thermal generators are located. To keep the system frequency stable, the total real power generation from the thermal generators must be exactly the same as the total net real power consumption. Hence, the real power set-point of all generators are determined by their capacity limits and load level. A power-flow solution is performed to obtain the voltage of nodes and current of branches by solving a set of circuit theory derived algebraic equations (power flow equations) of the network model. The solution of the algebraic equations is determined by loads, renewable generation and thermal generation. 


In our data generation procedure, we implement an iterative approach to create an uniform power-flow solution across the transmission system and multiple distribution systems. Each distribution system is represented as an equivalent load bus whose load value equals the sum of all nodes in the distribution system. We start with flat voltage level of 1.0 per-unit across the entire transmission system, change the loads and generator set-points according to the time-series data and solve standard power flow to obtain the voltage at the buses that has distribution system models. The voltage solutions are passed to each distribution system model as the voltage of the equivalent infinite source representing the transmission system. The distribution systems are then solved using the 'actual' voltage to obtain updated values of the real and reactive power consumption, as the load power consumption are related to the system voltage. The total powers of distribution systems are then passed back to the transmission system model and are used to update the power flow solution. This process is repeated until the voltage difference between two iterations converges to be less than a very small tolerance factor. The detailed simulation procedure is listed in Algorithm \ref{algo:ss} and the source code of our implementation using PSS/E and OpenDSS can be found in our GitHub repository.

\begin{algorithm}
\caption{Iterative Steady-State Power Flow Method for T+D Joint Simulation}
\label{algo:ss}
\begin{algorithmic}
\STATE Read transmission system case file in PSS/E with default parameters
\STATE Select hour from wind profile and set wind generators outputs
\STATE Select hour from load profile and scale load capacities
\STATE Initialize voltages all non-generator buses $V^0$ to 1 per-unit (flat start)
\STATE Initialize iteration counter $n = 1$
\WHILE{$||V^n - V^{n-1}||_2 > \epsilon$}
\FOR{$d$ in list of distribution circuits}
\STATE Read distribution system case files in OpenDSS with default parameters
\STATE Select hour from load and PV profiles and set net bus loads
\STATE Set source voltage to node voltage $V^{n-1}_d$ in transmission where the distribution circuit is located
\STATE Solve power flow and calculate total P and Q from all 3 phases
\STATE In transmission, adjust $P_d$ and $Q_d$ for the bus $d$ that has distribution circuit
\ENDFOR
\STATE Solve power flow using the updated P/Q values from distribution systems
\STATE Obtain all node voltages $V^n$ from results
\ENDWHILE
\STATE Simulation has converged, extract data to output file
\end{algorithmic}
\end{algorithm}

\subsubsection*{Millisecond-level Voltage, Current and Power Time Series}
\begin{figure}[hbpt!]
    \centering
    \includegraphics[width=0.85\textwidth]{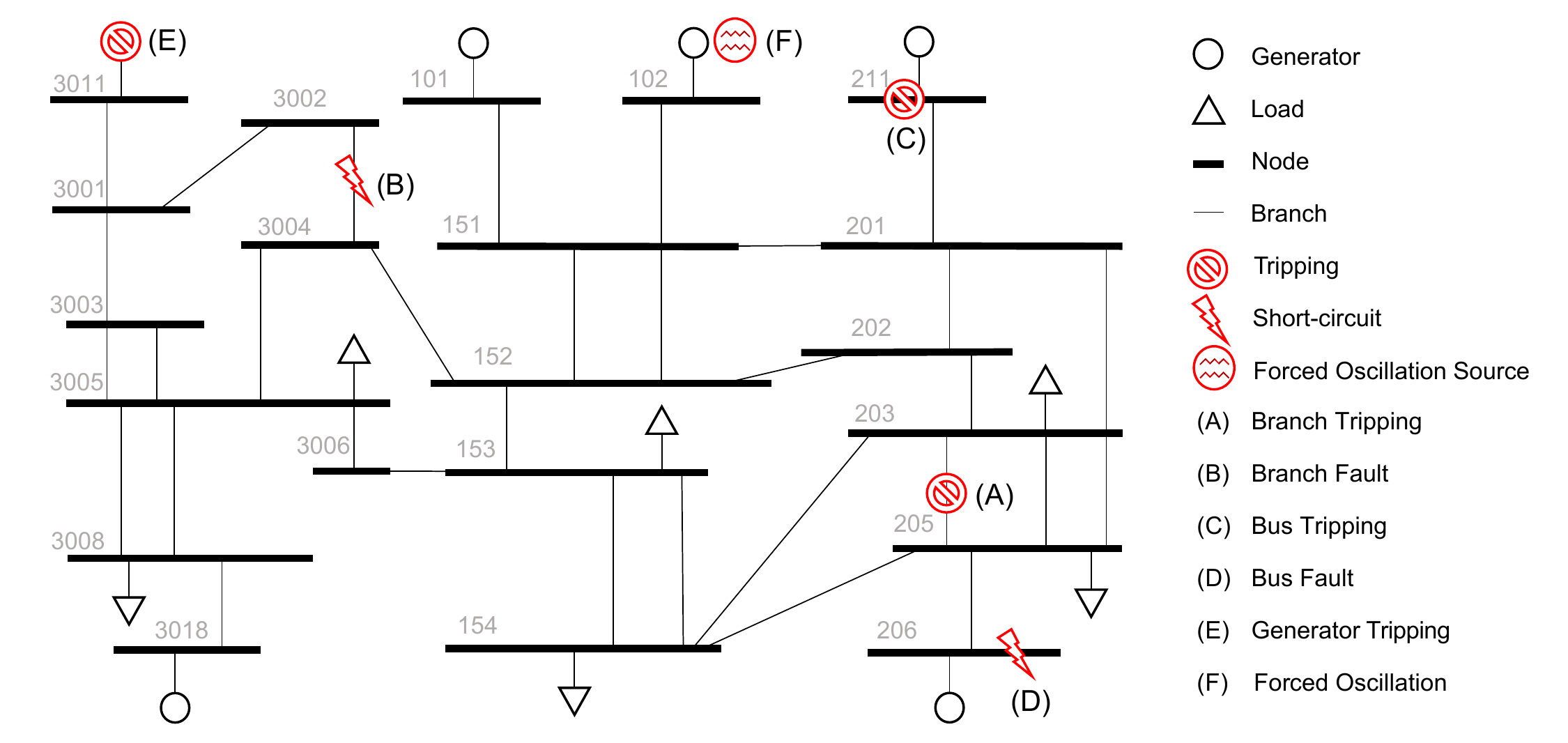}
    \caption{Visualization of different types of disturbances in the PSS/E 23-bus transmission system that induce the transient millisecond-level voltage, current and power measurements.}
    \label{fig:appendix:classification}
\end{figure}

In power systems there are many disturbances with different extent of types and severity, common ones include faults (short circuit between conductors or with the ground), unexpected equipment tripping and forced oscillation. These disturbances must be detected and handled as quickly as possible, or they may cause cascading failures and the damage can be out of control. Specifically, disturbances must be cleared within the Critical Clearing Time (CCT), usually at the order of 100 ms, after which the larger system will become unstable. Before the occurrence of a disturbance, the power system, at a larger time-scale, is assumed to be operating around a stable equilibrium point that is determined by the steady-state power flow solution related to generation and demand profiles. The initial conditions of dynamic components, including generators, wind turbines, solar PV systems and their many control devices, are determined by continuing random variations of the generation and demand capacity. During disturbances, the dynamics of the grid can be described by differential algebraic equations (DAEs). A transient simulation is essentially the process of solving those DAEs and obtaining timeseries of voltages, currents frequencies and other state variables. Fig. \ref{fig:cosimulation-flowchart} encapsulates the simulation mechanism, and Fig. \ref{fig:appendix:classification} illustrates several typical events of interested in co-simulation.

The approach to perform T+D co-simulation for transient events use a similar iterative algorithm to ensure that the data obtained from different systems and simulators are closely correlated with each other. The iteration process of exchanging voltage and power data between the bulk transmission system model and distribution system models is repeated for every time step in transient simulation. Beside the power system circuit and component models in PSS/E and OpenDSS, interver-interfaced solar PV dynamic models are implemented in Python and run separately. The simulation model developed for generating synthetic time series in \texttt{PSML} possesses the following features to obtain high-fidelity data: (i) The dynamic model of both the transmission and distribution systems are benchmark systems which are extensively used in power system research; (ii) we model the impact of deep penetration of renewables by incorporating detailed models of renewable generation and representative load and weather patterns in the U.S.; and (iii) the interaction between the transmission and the distribution systems in the fast time scale is modeled in the transient simulation, which is not captured in existing publicly available synthetic datasets, e.g., the oscillation dataset \cite{OLS}. 

\begin{algorithm}
\caption{Transient Event Simulation Method for T+D Joint Simulation}
\label{algo:tr}
\begin{algorithmic}
\STATE Select hour from wind, solar and load profiles
\STATE Solve steady-state power flow (Algorithm \ref{algo:ss}) as starting point
\STATE Create a disturbance with random on-time $T_{on}$, clear time $T_c$ and parameters
\STATE Create model objects for all PV generators and initialize with solar profile
\STATE Initialize transient simulation process in PSS/E using steady-state solution
\STATE Initialize time counter $t = 0$
\WHILE{$t < $ max step $T$}
\STATE Add/clear disturbance if $t == T_{on}$ or $T_c$
\FOR{$d$ in list of distribution circuits}
\STATE Read distribution system case files in OpenDSS with default parameters
\STATE Select hour from load and PV profiles and set net bus loads
\STATE Set source voltage to node voltage $V^{t}_d$ in transmission where the distribution circuit is located
\STATE Solve PV dynamic models and get net P/Q from PV generators
\STATE Solve power flow and calculate total P and Q from all 3 phases
\STATE In transmission, adjust $P_d^t$ and $Q_d^t$ for the bus $d$ that has distribution circuit
\ENDFOR
\STATE Advance OpenDSS transmission system transient simulation by one step
\ENDWHILE
\STATE Collect timeseries measurements and store to output file
\end{algorithmic}
\end{algorithm}

\subsection*{Machine Learning Benchmarks}
In this subsection, we select three types of grid-domain use cases for ML approaches. The use cases are (i) event detection, classification and localization, (ii) forecasting of renewable generation and load; and (iii) synthetic synchrophasor data generation. The reason that these use cases are selected is that they essentially can be formulated as classical ML problems which have been extensively studied during the past half-century. As a result, methods developed in the ML communities have great potential to provide solutions to these power grid use cases. Compared with conventional approaches heavily relying on grid physical models and network topology (line connectivity), such as the line outage detection algorithm \cite{4652583}, and the energy approach to forced oscillation localization \cite{maslennikov2017dissipating}, one attractive advantage of the ML-based approaches is that they do not require availability of information on grid physical model and topology. In addition, the sparsity, size, and scale of these time-series measurements provide a unique playground for the advancement of new ML methods. In what follows, we introduce the goal of each use case and present the benchmark of the performance of popular learning methods in terms of solving these three types of power system problems. We refer readers to our Github repository for more details on data structure and instructions on use cases.

\subsubsection*{Event Detection, Classification and Localization}
Renewable energy resources, such as wind/solar farms, are not as dispatchable as conventional fossil fuel generators due to their stochastic nature. As a result, those renewables introduce uncertain disturbances which may compromise the safe operation of the grid. Therefore, it is imperative for Independent System Operators (ISOs) to accurately recognize disturbances and perform corrective measures timely so as to ensure the safety of the grid. The health of the power grid is monitored by sensors such as synchrophasors/phasor measurement units (PMUs). These sensors stream time-stamped measurements to ISOs. Based on these streaming measurements, ISOs may have the following three questions: (i) When is an event happening; (ii) What type of event is happening? and (iii) Where is the source that caused the event? Answering these questions is critical to maintaining reliable operation of a power grid integrated with rich renewable energy resources.

The streaming measurements can be denoted by $X\in \mathbb{R}^{T\times M}$, where $T$ is the number of time stamps by now and $M$ is the number of measurements. \emph{Event detection} aims to answer the first ISO question by recognizing the disturbance occurrence once it takes place, hence a model $\mathcal{H}$ is learned to be able to identify the disturbance occurrence given sequence $X$, i.e., $\mathcal{H}: X\rightarrow \{0, 1\}$. Suppose the event takes place at time $\tau$: when $T<\tau$, the model is expected to be quiet without any alarms ($0$ predicted); when $T\geq\tau$, the model should alarm as soon as possible ($1$ predicted). 
\emph{Event classification} answers the second ISO question based on streaming sensor measurements. Given the measurement $X$, the objective of this task is to learn a model $\mathcal{F}$ that can classify the underlying event type $y$, i.e., 
$\mathcal{F}: X \rightarrow y$. In \texttt{PSML}, $y$ is a subset of disturbances $\mathcal{C}$ where
$\mathcal{C}:=$ $\{$\texttt{branch fault}, \texttt{branch tripping}, \texttt{bus fault}, \texttt{bus tripping}, \texttt{generator tripping}, \texttt{forced oscillation}$\}$.
\emph{Event localization} focuses on locating events (for branch fault, branch tripping, bus fault, bus tripping, generator tripping) or the root cause of events (for forced oscillations) by observing measurements. We are aimed at learning a model $\mathcal{G}$ that can map measurement $X$ to the bus(es) $z$ nearest to the events detected or the root cause of the events, i.e., $\mathcal{G}: X\rightarrow z$, where $z$ is a subset of buses $\mathcal{Z}$ in the entire system.
It is worth noting that compared with the size of the whole grid, the sensor coverage might be sparse in practice,  rendering the tasks of event detection, classification and localization more challenging.

We select the following representative benchmark algorithms for this task. We implement \emph{InceptionTime}, \emph{MC-DCNN} and \emph{ResNet} (with the help of sktime-dl~\cite{sktime-dl} package), and \emph{MLSTM-FCN}\cite{karim2019multivariate} in Tensorflow. We implemented all other methods by ourselves in Pytorch except \emph{TapNet}~\cite{zhang2020tapnet} and \emph{MiniRocket}~\cite{dempster2021minirocket}. For deep learning approaches, we use grid search to select general hyperparameters such as layer size, number of layers, normalization approach, etc.
\begin{itemize}
\item \textbf{Power domain}: Event localization is implemented by calculating the event signature of each PMU, where the event signature is estimated by several statistical parameters including Shannon entropy, standard deviation, range, mean difference and crest factor as introduced in the literature \cite{pandey2020real}. The PMU with the most dominant event signature indicates the location of the event.

\item \textbf{Traditional machine learning methods}: with generally good performance across different time series datasets, 1-nearest neighbor (1-NN) related approaches have been widely employed as standard benchmarks in the their corresponding benchmarks, e.g., UCR~\cite{dau2019ucr} and UEA~\cite{bagnall2018uea}. We consider three measures for sample distance computation: Euclidean and dynamic time warping with each feature dimension treated independently (\emph{DTW-i}) or dependently (\emph{DTW-d}). We also adopt \emph{MiniRocket}, where kernel transformations are firstly applied to time series followed by simple linear classifiers for time series classification~\cite{dempster2021minirocket}. 

\item \textbf{Convolutional Neural Networks}: benefiting from the deep convolutional neural networks (CNNs) and residual connections, vanilla and different variants of CNNs are consider to perform classification tasks: (i) InceptionTime~\cite{fawaz2020inceptiontime}: an ensemble of deep CNNs inspired by Inception-v4~\cite{szegedy2017inception} architecture; (ii) MLSTM-FCN~\cite{karim2019multivariate}: concatenation of LSTM of CNN for feature representation learning with an additional squeeze-and-excitation block for further performance improvement; (iii) ResNet~\cite{wang2017time}: adaptation of residual network from images~\cite{he2016deep} to multivariate time series; (iv) MC-DCNN~\cite{zheng2014time}: multi-channels deep CNNs where different temporal patterns are transformed firstly and then learned through separate convolutional layers; and (v) TapNet~\cite{zhang2020tapnet}: an attentional prototype network was incorporated into the convolutional layers to learn latent features for time series classification.   

\item \textbf{Other deep learning methods}: we also analyze the performance of general deep neural networks (i.e., fully-connected neural network) and specific time series deep models (i.e., RNN and its two variants~\cite{hochreiter1997long,chung2014empirical}, transformer~\cite{vaswani2017attention}) is evaluated.
\end{itemize}

For the training process, we randomly select 439 time-series from the millisecond transient PMU data as training samples and the remaining 110 time-series for testing. Each sequence has a metadata associated with the event type similar to the classification use case, i.e. \texttt{branch\_fault, branch\_trip, bus\_fault, bus\_trip, gen\_trip}. Each time-series has a sequence length of 960 observations, representing 4 seconds in the system recorded at 240 Hz. There are 91 dimensions for each time-series, including voltage, current and power measurements across the transmission system.
In the test process, with class imbalance in consideration, we adopt balanced accuracy to obtain performance of different classification methods for event classification and localization: \textit{balanced\_acc} = (\textit{sensitivity} + \textit{specificity}) / 2. Since both too early (false positive where no event happens yet) or too late (false negative where damage/cost accumulates along time without alarm or action) event detection are undesired, we leverage the macro-averaged mean absolute error, which considers the divergence between actual and predicted labels for ordinal regression on imbalanced datasets~\cite{baccianella2009evaluation}.

\subsubsection*{Load and Renewable Energy Forecasting}
\begin{figure}[hbpt!]
    \centering
    \includegraphics[width=1.0\textwidth]{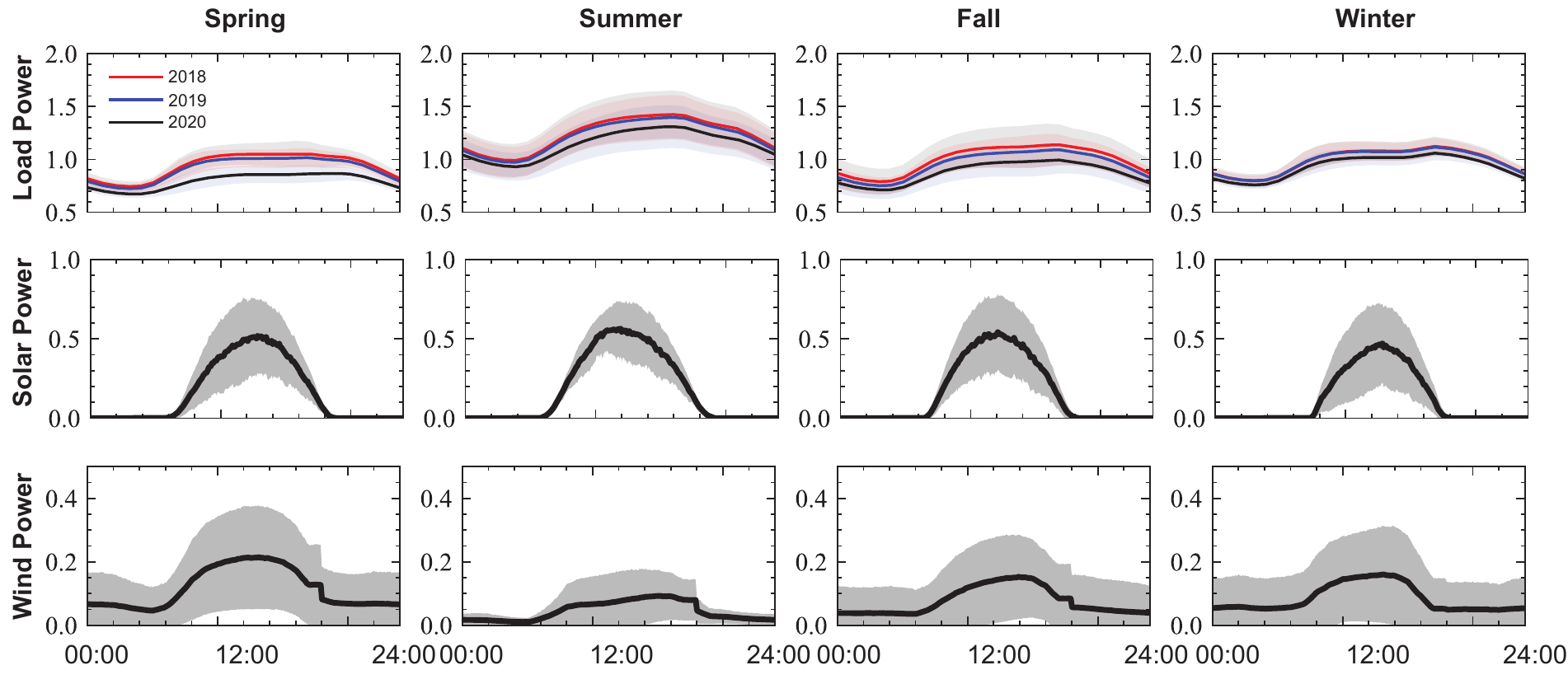}
    \caption{Illustration of daily load, solar and wind power profiles with 1-minute resolution sampled in Houston, capturing seasonal disparity, strong variation of renewables, and unprecedented load drop during pandemic. The solid lines represent the average and shaded areas represent standard deviation.}
    \label{fig:forecasting}
\end{figure}

The ultimate goal of the power grid is to balance generation and load. Today, this is mostly achieved by load forecasting and generation scheduling based on the forecast before real-time operation. In  real-time operation, the relatively small mismatch between scheduled generation and actual load is compensated by dispatchable generation units that can respond quickly, i.e., the spinning reserve. The spinning reserve may rely on fossil fuel and incur high operational costs. However, with increasing renewable integration, this  operational paradigm is not feasible without accurate renewable and load forecasting. Renewable generations, e.g., wind/solar farms, have their maximum output apparent power determined by weather, which cannot be actively increased (but can be curtailed), due to their stochastic and volatile nature. A poor forecast of renewable generation therefore leads to a large amount of expensive, fossil fuel-based spinning reserve being committed. Compounding the challenge, loads will become less predictable in the future grid, due stochastic loads like electric vehicles, and small-probability yet high-impact events, e.g., COVID-19 pandemic (Fig. \ref{fig:forecasting}) and the Texas winter storm in 2021. Therefore, accurate forecasting of renewable generation and load is critical to support reliable operation of the future grid. 

We focus on the following two important subtasks: {(i) Point Forecast (PF):} given a time sequence in the past $k$ time steps, the current time $t$ and the forecasting horizon $\tau$, we have the past targets $y_{t-k:t}$, the past observations $x_{t-k:t}$, as well as the past, current and future known variables $u_{t-k:t+\tau}$ (e.g., date, holiday). We aim to predict the targets $\tau$ time steps ahead: $\hat{y}_{t+\tau}=r(y_{t-k:t}, x_{t-k:t}, u_{t-k:t+\tau})$. {(ii) Prediction Interval (PI):} Uncertainty quantification can provide more reference information for decision-making and has received growing research and industrial interests these years~\cite{khosravi2011comprehensive,makridakis2020m4}. For high-quality uncertainty quantification, we would also like to obtain the prediction interval, $[\hat{y}^L, \hat{y}^H]$, to cover the ground truth $y$ with at least the expected tolerance probability, i.e., $p=0.95$ in our load and renewable energy forecasting task.

We select the following representative benchmark algorithms for this task. We implement the time-series models with statsmodels~\cite{seabold2010statsmodels}, the traditional machine learning models with sklearn~\cite{sklearn_api}, \emph{N-BEATS}~\cite{Oreshkin2020:N-BEATS}, \emph{WaveNet}~\cite{oord2016wavenet}, \emph{TCN}~\cite{BaiTCN2018}, \emph{LSTNet}~\cite{lai2018modeling}, \emph{DeepAR}~\cite{salinas2020deepar}, \emph{Informer}~\cite{haoyietal-informer-2021} and \emph{Neural ODE}~\cite{chen2018neural} with codes published officially (or unofficially), and all other deep learning approaches by ourselves in Pytorch. For deep learning approaches, we use grid search to select general hyperparameters such as layer size, number of layers, normalization approach, etc.

\begin{itemize}
\item \textbf{Time-series models}: besides the \emph{naive} method takes the current value directly as the prediction, we also consider autoregressive integrated moving average (ARIMA) and exponential smoothing (ETS).
\item \textbf{Traditional machine learning methods}: we select the top four widely used machine learning methods in load and renewable energy forecasting literature~\cite{ahmad2020review}: support vector regression (SVR), random forest (RF), gradient boosted decision trees (GBDT), and linear regression (LR).
\item \textbf{Multilayer perceptron}: besides fully-connected neural newotks (FNN) and extreme learning machines (ELM), we also list performance of N-BEATS, a deep stacked neural architecture based on backward and forward residual links, which outperformed the winner of M4 competition~\cite{oreshkin2019n}.
\item \textbf{Convolutional Neural Networks}: we study performance of vanilla CNN, WaveNet~\cite{oord2016wavenet} composed of dilated causal convolutions for audio generation, Temporal convolutional neural networks (TCN~\cite{lea2017temporal}) with additional residual blocks. 
\item \textbf{Recurrent Neural Networks}: both basic recurrent neural networks (vanilla RNN, LSTM~\cite{hochreiter1997long} and GRU~\cite{chung2014empirical}) and advanced variants are studied: LSTNet~\cite{lai2018modeling} with patterns extracted from convolutional layers and fed to recurrent neural networks, DeepAR~\cite{salinas2020deepar} with output from recurrent neural networks as likelihood parameters for probabilistic forecasting.
\item \textbf{Transformer-based}: we list performance from vanilla transformer~\cite{vaswani2017attention} and its variant, informer~\cite{zhou2021informer}, with self-attention distilling and generative style decoder for long sequence forecasting.
\item \textbf{Neural ODE}: motivated by the Euler discretization of continuous transformations in residual networks and recurrent neural network decoders, Neural ODEs parameterize the derivative of hidden state using a neural network and compute the network output with a differential equation solver~\cite{chen2018neural}
\end{itemize}

Given load and renewable energy data recorded from 66 locations in minute-level, we split the sequence from each location according to years firstly and have three cases: (i) use data from Jan to Nov in 2018 for training and Dec in 2018 for testing; (ii) use data from Jan, 2018 to Nov, 2019 for training and Dec in 2019 for testing; (iii) use data from Jan, 2019 to Nov, 2020 for training and Dec in 2020 for testing. Noted that we adopt the rolling strategy during testing, so that testing data before current time step is observable for model forecasting. 
In the test process, we adopt three commonly leveraged metrics in load and renewable energy forecasting literature~\cite{ahmad2020review}: root mean square error (RMSE), mean absolute error (MAE) and mean absolute percentage error (MAPE). Following practice in the M4 competition~\cite{makridakis2020m4}, the performance of generated point intervals is evaluated using the mean interval score (MIS)~\cite{gneiting2007strictly}:
\begin{equation}
    \text{MIS}=\frac{1}{N}\sum_{i=1}^N(\hat{y}^u_i-\hat{y}^l_i)+\frac{2}{a}(\hat{y}^l_i-y_i)\mathcal{I}(y_i<\hat{y}^l_i)+\frac{2}{a}(y_i-\hat{y}^u_i)\mathcal{I}(y_i>\hat{y}^u_i)
\end{equation}
where $N$ is the number of instances for each prediction horizon, $\mathcal{I}$ is the indicator function with value $1$ when the inequality holds and $0$ otherwise, and $a=0.05$ for $95\%$ prediction intervals generation.

\subsubsection*{Synthetic Time-series Generation}
A major hurdle in applying deep learning models to power system problems is usually the lack of sufficient and high-quality datasets for training, as it is well-known that more eventful data usually lead to better classification performance \cite{gaing2004wavelet, okumus2018power, 9361704}. 
The accessibility of real-world power grid PMU measurement data is limited due to the regulation CEII \cite{ceii} for national security and sensitivity concerns. While researchers recently have contributed to the creation of large-scale synthetic simulation models \cite{birchfield2016grid} for analysis \cite{BTE1,wu2021open}, there are always gaps between simulation models and real-world systems and the unique values of real-world PMU time series data cannot be exploited for research purposes. It is therefore critical to investigating methods for synthesizing power system datasets that follow the same properties of the real system data while 
complying with physical laws for the network and its underlying dynamic behaviors. 

This task involves multi-channel time series generation, for which the training data are disturbance-induced dynamic voltage, current and power measurements across power grids. The expected outputs are dynamic voltage, current and power measurements that preserve certain dynamic patterns and physical laws. The key challenges that distinguish this task from normal image generation are: (i) multi-channel time series are governed by unknown algebraic and differential equations derived from physical laws, and (ii) dynamic time series incorporate discrete disturbance events. Our evaluations are carried out over simulated voltage, current and power data from \texttt{PSML}.

We select the following representative benchmark algorithms for this task. All models are trained with a fixed hidden dimensionality of 256, a fixed number of two or three layers for recurrent networks, and a tuned dropout ratio $\in$ \{0.0, 0.5\}.
\begin{itemize}
    \item \textbf{NaiveWGAN}: we show the performance of a naive GAN architecture (MLP generator and discriminator) with the Wasserstein loss \cite{arjovsky2017wasserstein}.
    \item \textbf{RCGAN}: a conditional recurrent GAN architecture \cite{esteban2017real} is tested  that leverage recurrent generator and discriminator and conditioned on auxiliary information.
    \item \textbf{COT-GAN}: we test a recurrent GAN trained with a  Causal Optimal Transport (COT) loss suitable for learning time dependent data distributions \cite{xu2020cot}.
    \item \textbf{TimeGAN}: we list the performance of a recurrent GAN architecture that combines unsupervised GAN learning with a supervised teacher-forcing component in the loss function \cite{yoon2019time}.
    \item \textbf{DoppelGANger}: we test a state-of-the-art GAN architecture \cite{lin2020using} that leverages two generators and discriminators to first generate auxiliary metadata before generating the time-series.
\end{itemize}

For training and testing the models, we set the first 400 millisecond transient time-series as training samples and the next 150 time-series for testing, where each time-series has a sequence length of 960 observations, representing 4 seconds in the system recorded at 240 Hz, and has 91 dimensions.
As the task is to synthesize multiple realistic-looking PMU streams that respect the physical constraints from real PMU streams, we define the following metrics to assess the quality of generated data: (i) \textit{Fidelity}: samples should be indistinguishable from the real data. We train a post-hoc time-series classification model (by optimizing a 2-layer LSTM) to distinguish between sequences from the original and generated datasets and report the error, and (ii) \textit{Diversity}: samples should be distributed to cover the real data. We apply PCA analyses on both the original and synthetic datasets (flattening the temporal dimension) and visualize how closely the distributions are in 2D space.

\section*{Data Records}
The dataset is hosted on Zenodo~\cite{PSML_dataset}. The folders in the dataset are organized by data type, including minute-level load and renewable data, minute-level PMU measurements data, and millisecond-level PMU measurements data.

In the folder of the minute-level load and renewable data, we store the data in the CSV files by zonal location.
\begin{itemize}
        \item \textbf{File} ISO\_zone\_\#.csv: for example, \emph{CAISO\_zone\_1.csv} is a CSV file containing munute-level laod, renewable and weather data from 2018 to 2020 in the zone 1 of CAISO.
        \begin{itemize}
            \item \textbf{Field} \emph{time}: time of minute resolution.
            \item \textbf{Field} \emph{load\_power}: time of minute resolution.
            \item \textbf{Field} \emph{wind\_power}: time of minute resolution.
            \item \textbf{Field} \emph{solar\_power}: time of minute resolution.
            \item \textbf{Field} \emph{DHI}: diffuse horizontal irradiance.
            \item \textbf{Field} \emph{DNI}: direct normal irradiance.
            \item \textbf{Field} \emph{GHI}: global horizontal irradiance.
            \item \textbf{Field} \emph{Dew Point}: dew point in degree Celsius.
            \item \textbf{Field} \emph{Solar Zeinth Angle}: angle in degree between the sun's rays and the vertical direction.
            \item \textbf{Field} \emph{Wind Speed}: wind speed in meter per second.
            \item \textbf{Field} \emph{Relative Humidity}: relative humidity in percentage.
            \item \textbf{Field} \emph{Temperature}: temperature in degree Celsius.
        \end{itemize}
\end{itemize}

In the folder of the minute-level PMU measurements data, we store the data in the subfolders by scenario setting. Each subfolders contains a TXT file and a CSV file that respectively store the metadata and simulation measurements data.
\begin{itemize}
    \item \textbf{Folder} case \#: for example, \emph{case 0} is a folder corresponding to the scenario setting \#0.
    \begin{itemize}
        \item \textbf{File} pf\_input\_\#.txt: the TXT file contains the metadata of the selected load and renewable data for the simulation.
        \item \textbf{File} pf\_result\_\#.csv: the CSV file contains the voltage at buses and power on branches in the transmission system via T+D simulation.
        \begin{itemize}
            \item \textbf{Field} \emph{time}: time of minute resolution.
            \item \textbf{Field} \emph{Vm\_\#}: voltage magnitude in per unit at the bus \#.
            \item \textbf{Field} \emph{Va\_\#}: voltage angle in rad at the bus \#.
            \item \textbf{Field} \emph{P\_\#\_\#\_\#}: for example, \emph{P\_3\_4\_1} means the active power in the \#1 branch from the bus 3 to 4.
            \item \textbf{Field} \emph{Q\_\#\_\#\_\#}: for example, \emph{Q\_5\_20\_1} means the reactive power in the \#1 branch from the bus 5 to 20.
        \end{itemize}
    \end{itemize}
\end{itemize}

In the folder of the millisecond-level PMU measurements data, we seperate the data into two folders by oscillation type, namely forced oscillation and natural oscillation. In each folder, the data are organized in the following way.
\begin{itemize}
    \item \textbf{Folder} row\_\#: for example, \emph{row\_0} is a folder corresponding to the disturbance scenario \#0.
    \begin{itemize}
        \item \textbf{File} dist.csv: the CSV file contains the three-phased voltage at nodes in the distribution system.
        \begin{itemize}
            \item \textbf{Field} \emph{Time(s)}: time of millisecond resolution.
            \item \textbf{Field} \emph{\#.\#.\#}: for example, \emph{3005.633.1} means the per-unit voltage magnitude of the phase A at the bus 633 of the distribution grid, the one connecting to the bus 3005 in the transmission system..
        \end{itemize}
        \item \textbf{File} trans.csv: the CSV file contains the voltage at buses and power on branches in the transmission system.
        \begin{itemize}
            \item \textbf{Field} \emph{Time(s)}: time of millisecond resolution.
            \item \textbf{Field} \emph{VOLT \#}: voltage magnitude in per unit at the bus \#.
            \item \textbf{Field} \emph{POWR \# TO \# CKT \#}: for example, \emph{POWR 151 TO 152 CKT '1 '} means the active power transferring in the \#1 branch from the bus 151 to 152.
            \item \textbf{Field} \emph{VARS \# TO \# CKT \#}: for example, \emph{VARS 151 TO 152 CKT '1 '} means the reactive power transferring in the \#1 branch from the bus 151 to 152.
        \end{itemize}
    \end{itemize}
\end{itemize}

Besides, the algorithm codes are available on Github~\cite{PSML_github_code}. In the folder \emph{Code}, we organize the codes for algorithm reproduction as follows. Please refer to the Usage Notes for the details of installation, package usage and navigation.
\begin{itemize}
    \item \textbf{File} \emph{dataloader.py}: Pytorch data loaders with both data processing and splitting included.
    \item \textbf{File} \emph{evaluator.py}: evaluators to support fair comparison among different approaches.
    \item \textbf{Folder} \emph{BenchmarkModel}: .
    \begin{itemize}
        \item \textbf{Folder} \emph{EventClassification}: the folder contains all files for the event classification task.
        \begin{itemize}
            \item \textbf{Folder} \emph{configs}: the folder contains all trained model configurations (YAML files).
                \begin{itemize}
                    \item \textbf{File} \emph{\#.yaml}: for example, \emph{RNN.yaml} stores the model parameters of the RNN model.
                \end{itemize}
            \item \textbf{Folder} \emph{models}: the folder contains all model codes (PY files).
                \begin{itemize}
                    \item \textbf{File} \emph{\#.py}: for example, \emph{RNN.py} implements the RNN model.
                \end{itemize}
            \item \textbf{File} \emph{evaluating.py}: codes to evaluate the trained benchmark models for event classification.
            \item \textbf{File} \emph{processing.py}: codes to process the millisecond-level measurements data.
            \item \textbf{File} \emph{requirements.txt}: the required Python packages for model use.
        \end{itemize}
        \item \textbf{Folder} \emph{LoadForecasting}: the folder contains all files for the forecasting task.
        \begin{itemize}
            \item \textbf{Folder} \emph{configs}: the folder contains all trained model configurations (YAML files).
                \begin{itemize}
                    \item \textbf{File} \emph{\#.yaml}: for example, \emph{FNN.yaml} stores the model parameters of the FNN model.
                \end{itemize}
            \item \textbf{Folder} \emph{models}: the folder contains all model codes (PY files).
                \begin{itemize}
                    \item \textbf{File} \emph{\#.py}: for example, \emph{FNN.py} implements the FNN model.
                \end{itemize}
            \item \textbf{File} \emph{evaluating.py}: the codes to evaluate the trained banchmark models for forecasting.
            \item \textbf{File} \emph{processing.py}: the codes to process the minute-level load and renewable data.
            \item \textbf{File} \emph{requirements.txt}: the required Python packages for model use.
        \end{itemize}
        \item \textbf{Folder} \emph{SyntheticDataGeneration}: the folder contains all files for the synthetic data generation task.
        \begin{itemize}
            \item \textbf{Folder} \emph{\#\#}: for example, \emph{RGAN} contains all files for the RGAN model.
                \begin{itemize}
                    \item \textbf{File} \emph{requirements.txt}: the required Python packages for model use.
                    \item \textbf{Other files}: other files contained in the folder depends on the model, of which the usage are instructed in the README file.
                \end{itemize}
            \item \textbf{File} \emph{evaluating.py}: the codes to evaluate the trained banchmark models for forecasting.
            \item \textbf{File} \emph{processing.py}: the codes to process the minute-level load and renewable data.
        \end{itemize}
    \end{itemize}
\end{itemize}

\section*{Technical Validation}
\subsection*{Dataset Explorations}
\subsubsection*{Load and Renewable Time Series}
The zonal load and weather data are collected from multiple real-world sources, of which the fidelity is guaranteed by manual data quality control. The associated physical models for renewable estimation are commonly used in the power system. We illustrate daily load and renewable power profiles in Fig. \ref{fig:forecasting}, which shows seasonal disparity and strong variation of renewable energy. We can also observe a significant load reduction during the COVID-19 pandemic, which is valuable for investigating the impacts of unprecedented events on the energy sector. 

\subsubsection*{Minute-level Voltage, Current and Power Time Series.}
\begin{figure}[hbpt!]
\centering
\includegraphics[width=0.5\columnwidth]{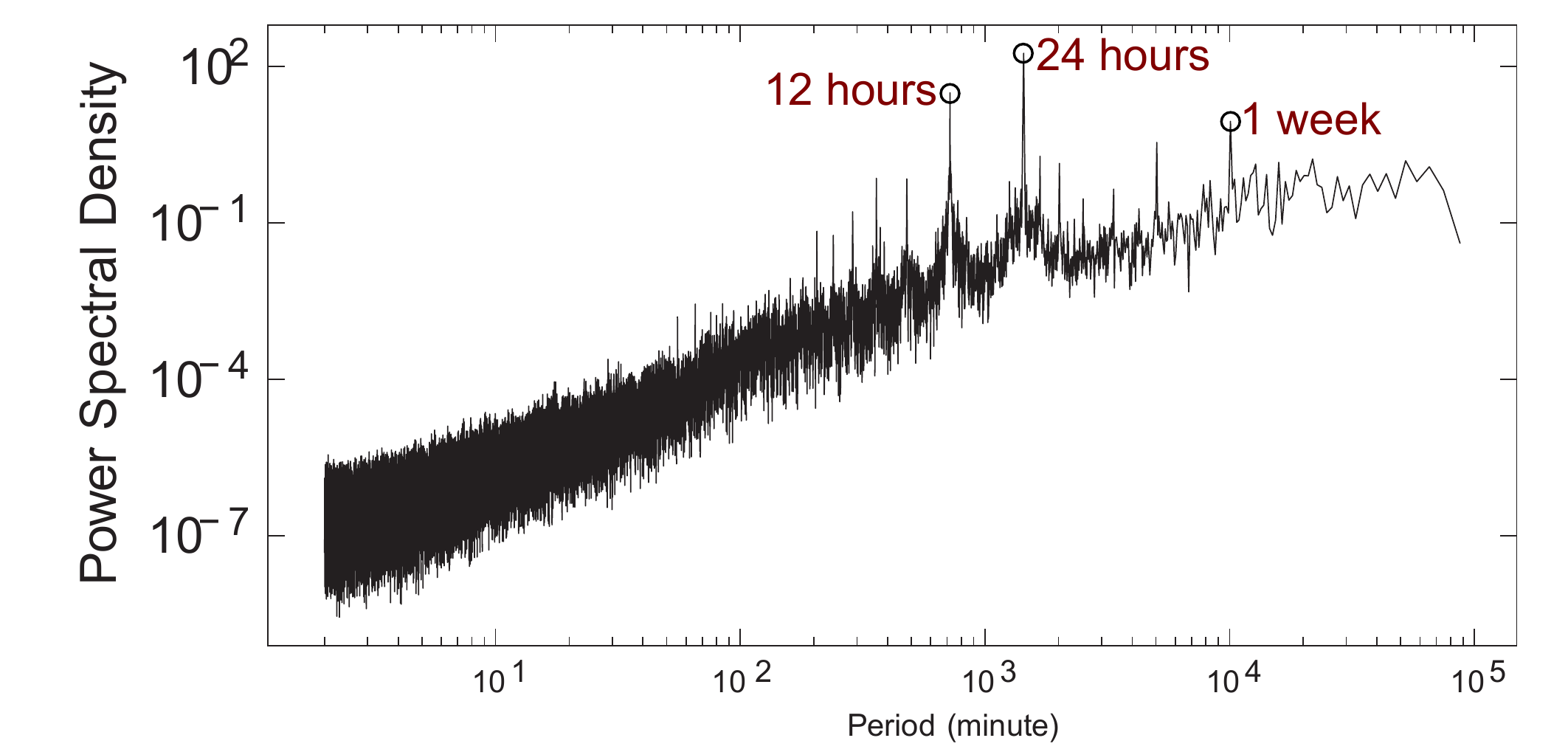}
\caption{Power spectral density of minute-level PMU measurements of 1 year. It demonstrates the data fidelity by the high power density at the periods of 12 hours, 24 hours and 1 week, which is consistent with our prior observations in other real-world PMU dataset.}
\label{fig:PSD}
\end{figure}

To demonstrate the data fidelity, we illustrate the power spectral density analysis that quantifies the periodicity on one selected 1-year-long minute-level synthetic PMU measurement data. Specifically, we calculate the power spectral density of the voltage angle at each transmission system bus. Fig.~\ref{fig:PSD} shows the average power spectral density, where the highest power density appear at the period of 12 hours, 24 hours and 1 week. It is consistent with our prior observation in other real-world PMU dataset that the major periodic patterns are derived from the loads that mainly possess periods of 12 hours, 24 hours and 1 week.

\subsubsection*{Millisecond-level Voltage, Current and Power Time Series.}
\begin{figure}[hbpt!]
    \centering
    \includegraphics[width=1.0\textwidth]{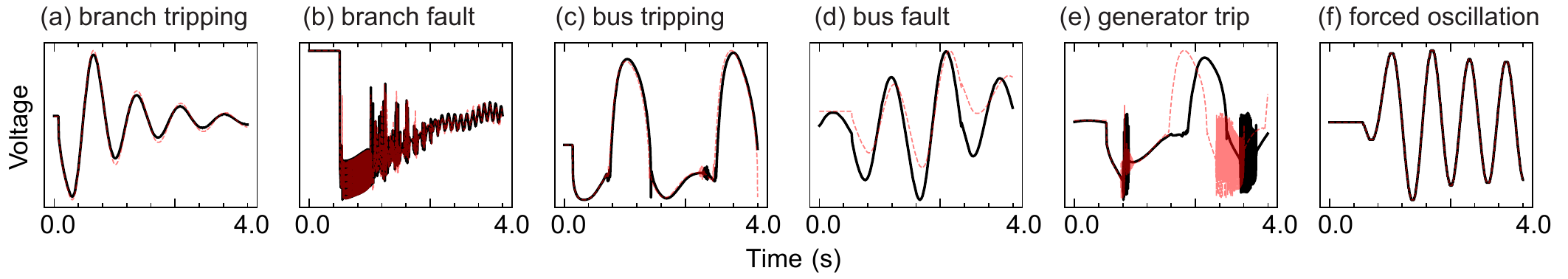}
    \caption{Illustration of voltage magnitude time series of millisecond resolution in cases of different types of disturbances, where the black solid profiles are critical features captured by the novel T+D co-simulation, some of which are missed by the red dashed lines obtained conventional transmission system simulation alone.}
    \label{fig:classification}
    \vspace{-0.4cm}
\end{figure}

\begin{figure}[hbpt!]
    \centering
    \includegraphics[width=0.8\textwidth]{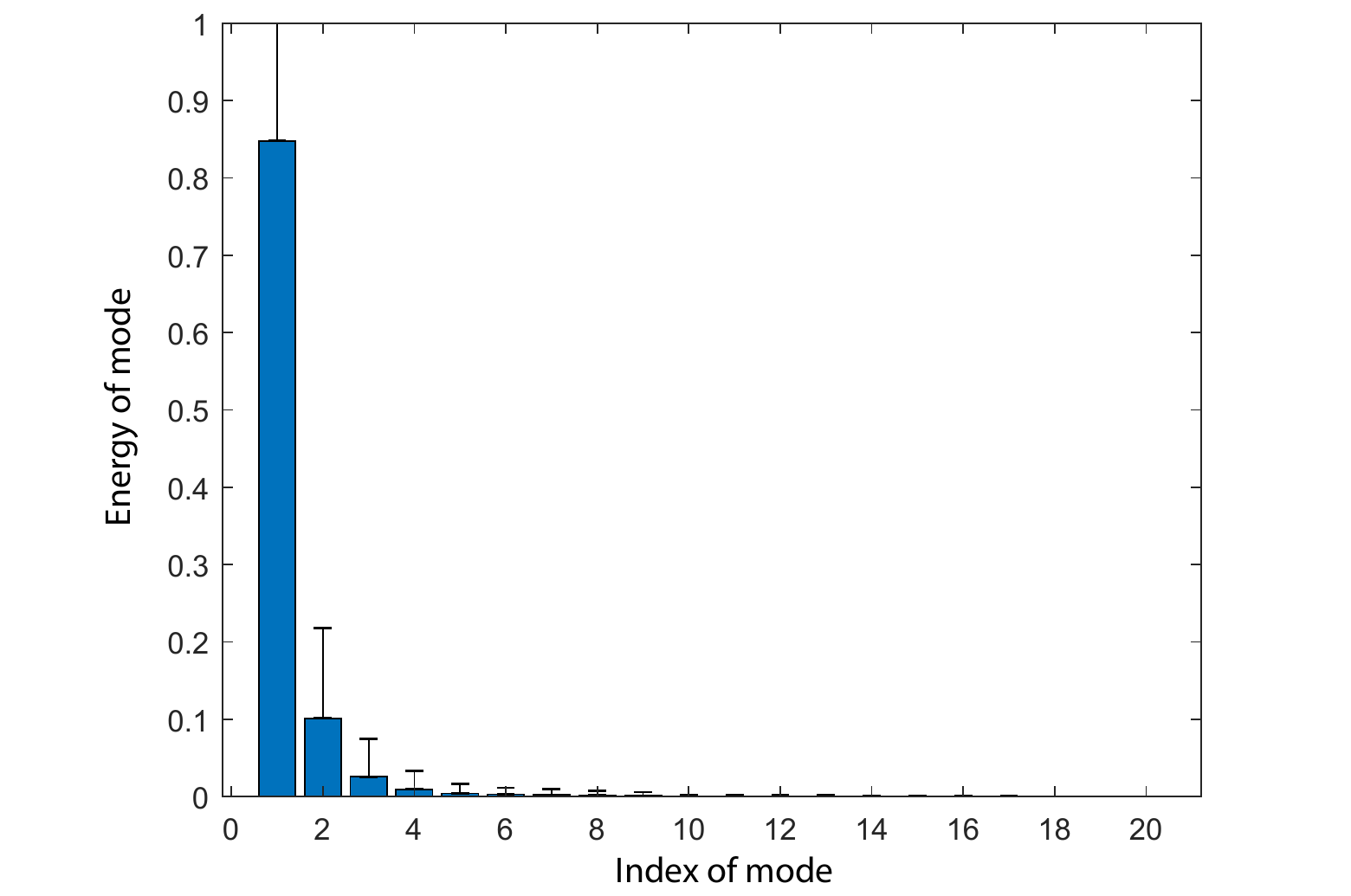}
    \caption{Modal analysis of millisecond-level PMU measurements, showing limited number of dominant modes of high energy.}
    \label{fig:modal_analysis}
    \vspace{-0.4cm}
\end{figure}

Fig. \ref{fig:classification} visualizes some typical events obtained from the co-simulation, compared with the profiles by traditional simulation. It shows that the co-simulation method reveals more details.
We further perform modal analysis on the generated power time series to demonstrate the fidelity. We observe in Fig.~\ref{fig:modal_analysis} that the generated power time series data possess only few dominant modes of high energy, which match our prior knowledge.

\subsection*{Machine Learning Benchmark Evaluation}
\subsubsection*{Event Detection, Classification and Localization}
As listed in Table~\ref{tab:classification_performance}, we evaluate methods from four main categories: (i) PMU score from the power community, (ii) standard classification benchmarks based on 1-nearest neighbor (1-NN), (iii) convolutional neural networks and (iv) other popular deep learning methods.
For both event classification and detection, we observe in general that approaches composed of convolutional neural networks perform much better than 1-NN based standard benchmarks and other deep learning approaches. However, to localize the event across the grid, the majority of deep learning approaches fail to achieve competitive performance as 1-NN approaches: only \emph{MC-DCNN} reaches above $0.40$ balanced accuracy besides 1-NN Euclidean, 1-NN DTW-i and 1-NN DTW-d. 
In analyzing PMU measurements from the transmission system, we attribute the success of convolutional neural networks to their explicit spatial correlation modeling, where the voltage and current evolve along time according to both the external oscillation events and the inherent network connectivity.

Based on the above observations, we expect more accurate classifications can be obtained by proposing more powerful deep learning methods from but not limited to the following directions. We hope that these interesting directions will motivate the ML community in addressing the challenging problems of event detection, classification and localization for dynamical systems that exhibit tight multi-scale spatio-temporal coupling.
\begin{itemize}
    \item \textbf{Graph neural networks with both spatial and temporal dependencies:} event localization is a great challenge when only temporal dependencies are modeled in deep learning approaches. Given bus locations and their connectivity, graph neural networks may be promising in modeling spatial dependencies and locating the actual event bus.
    \item \textbf{Incorporating contrastive learning into representation learning:} by comparing two instances ($x_i$ and $x_j$) rather than learning the mapping from $X$ to $y$, 1-NN based approaches outperform deep learning approaches in event localization. Recently, representation learning based on comparing three instances (that is, one anchor, positive, and negative sample in triplet loss) or two instances (e.g., two similar samples by transformation or same annotated label in Siamese approaches) shows effectiveness in capturing underlying patterns and further benefits downstream tasks in domains like computer vision~\cite{taigman2014deepface}, reinforcement learning~\cite{grill2020bootstrap}, etc. 
\end{itemize}

\subsubsection*{Load and Renewable Energy Forecasting}
\begin{figure}[hbpt!]
\centering
\includegraphics[width=.7\columnwidth]{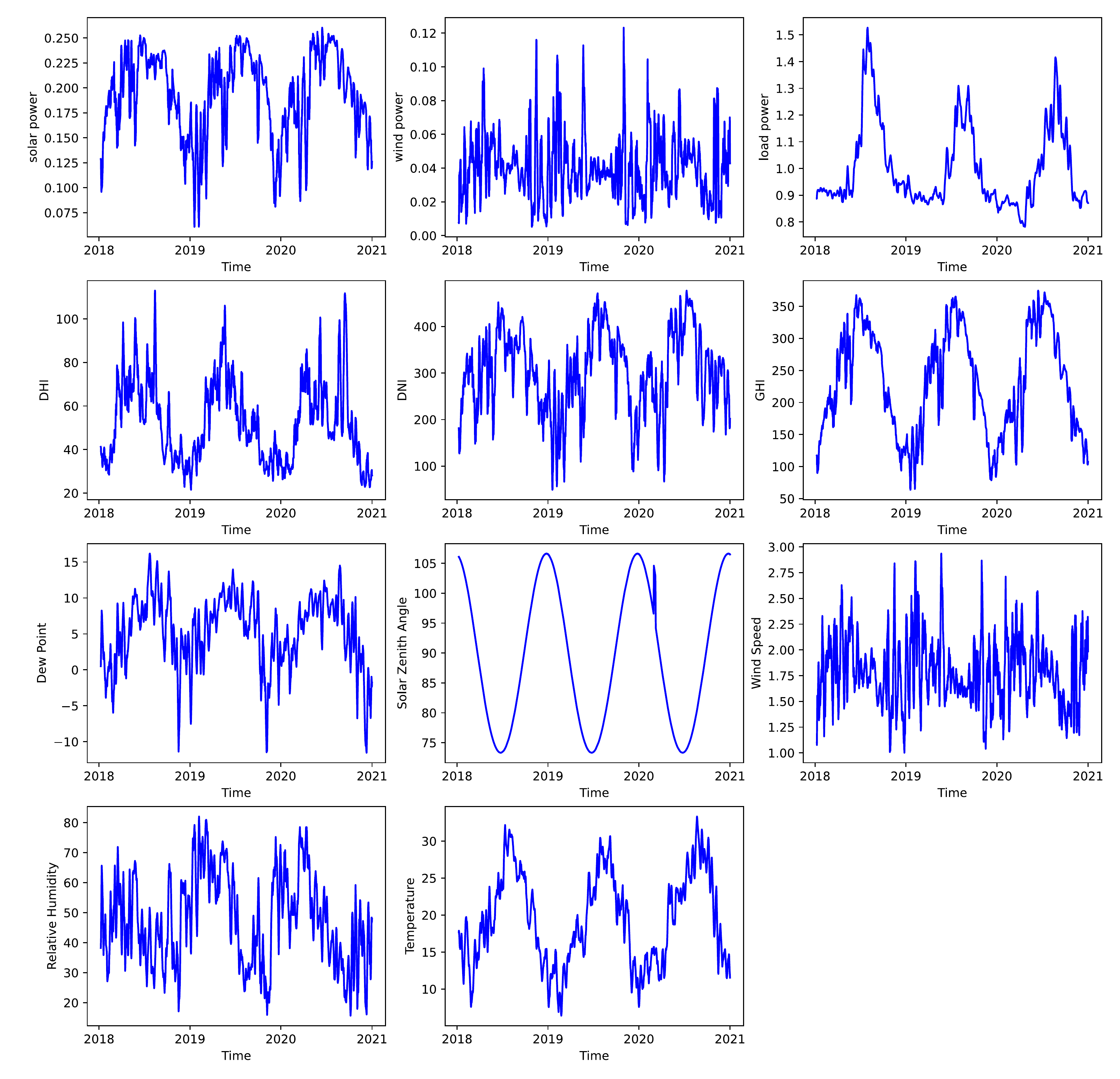}
\caption{Visualization of feature trends in three years from one sampled location of the forecasting task.}
\label{fig:forecasting_dataset_visualization}
\end{figure}

We list the performance of different forecasting methods on 1-hour ahead load forecasting in years from 2018 to 2020 in Table~\ref{tab:forecasting_performance_load_t+60}, while we list the performance on all the forecasting tasks on our Github. 
For short-term load forecasting, exponential smoothing outperforms the other benchmarks both in point forecast and prediction interval, while deep learning approaches fail to achieve competitive performance on par with time series models and traditional machine learning methods. Similar observations can be discovered in both short- and long-term forecasting of wind and solar. 
As visualized in Fig.~\ref{fig:forecasting_dataset_visualization}, we can observe strong periodicity in both the observational features (such as wind speed, relative humidity, and temperature) as well as target features (such as solar power, wind power, and load) from the year 2018 to year 2020. However, deep learning approaches fail to capture such significant trends for accurate future forecasting due to their limited memorizing capabilities. For extremely long time series, it's difficult to efficiently extract and leverage useful past time steps without tedious feature engineering~\cite{wang2019deep}. Taking our forecasting task as an example: simply enlarging the scope of considered historical data to cover information from the previous $1440$-th, $10,080$-th, $43,200$-th time step in our minute-level data for the reference of the same time in the last day, same weekday in the previous week, and same day in the previous month, is both time-consuming during processing crowded useless information and will deteriorate forecasting performance in the end.

Accordingly, we suggest the following directions for further exploration. We believe that this task can motivate the development of novel deep learning based forecasting algorithms that take into account long-term memory and exploit spatio-temporal patterns in the data.
\begin{itemize}
    \item \textbf{Memory network to remember and utilize past history efficiently:} approaches such as memory networks~\cite{weston2014memory} could be potential solutions to identify key information from long histories for real-time forecasting.
    \item \textbf{Cross-learning:} using information from multiple series to predict individual ones has shown promising results in top approaches of past Kaggle competitions~\cite{bojer2021kaggle}. Taking into account load and renewable energy time series from other locations, such as nearby locations in the same time zone or with similar social and economic patterns, could potentially enhance forecasting accuracy.
\end{itemize}

\subsubsection*{Synthetic Time-series Generation}
\begin{figure}[hbpt!]
\centering
\includegraphics[width=\columnwidth]{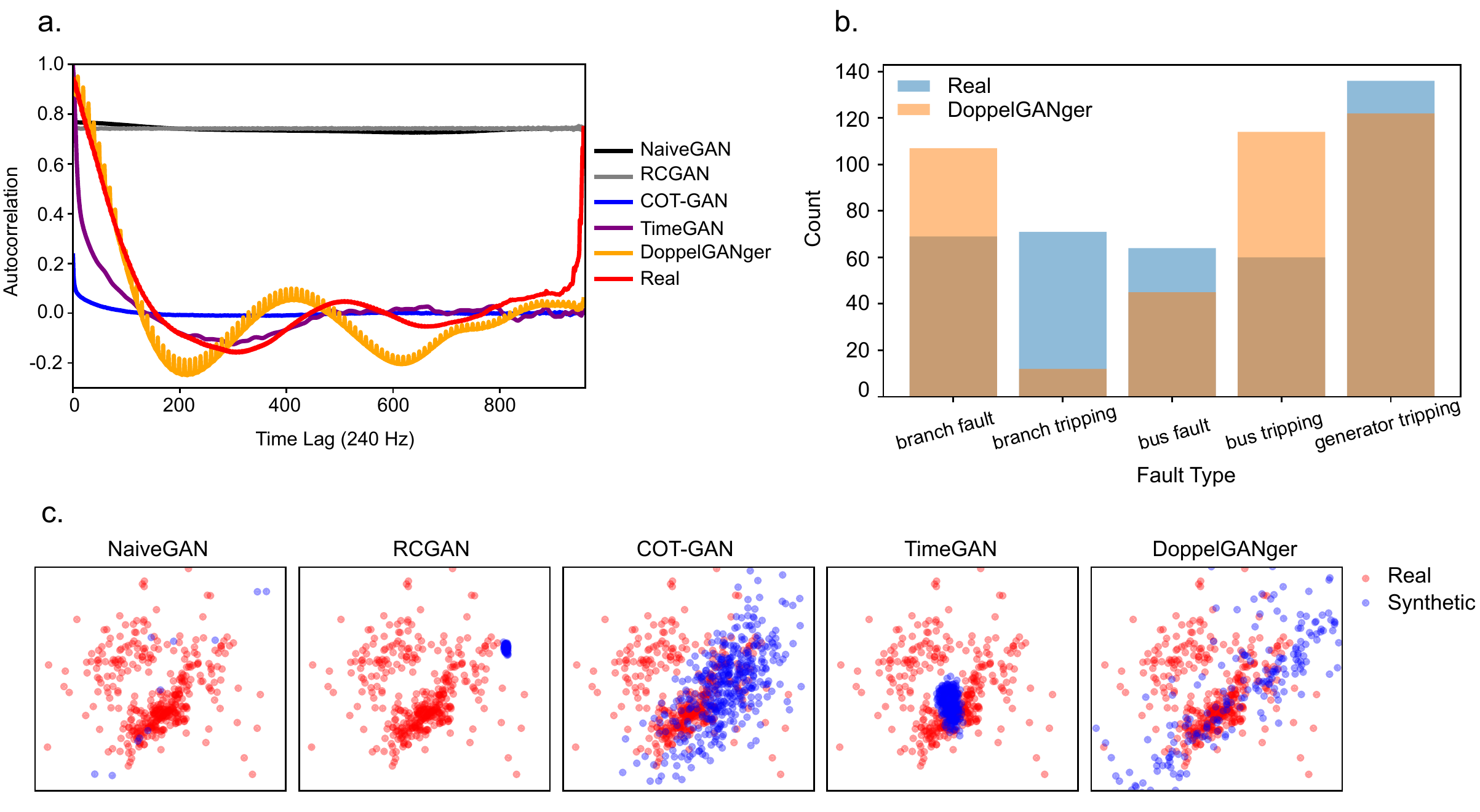}
\caption{Performance of generative methods. a. Autocorrelation on PMU datasets for all generative methods. b. Distribution of generated faults for DoppelGANger. c. PCA coverage evaluation on millisecond-level eventful PMU datasets.}
\label{fig:performance_aggregation}
\end{figure}

We observe that current SOTA time-series generation methods cannot properly capture the necessary characteristics from PMU data to generate realistic time-series. 
This is reflected in the fidelity metrics in Table~\ref{tab:fidelity},
where a post-hoc 2-layer LSTM can easily separate real vs. generated samples. 
Time-GAN \cite{yoon2019time} and COT-GAN \cite{xu2020cot} are current SOTA methods published in NeurIPS 19 and 20 respectively and they both achieve relatively low auto-correlation and cross-correlation error compared to real data (Fig.~\ref{fig:performance_aggregation}-a). 
In addition, we apply PCA analyses on both the original and synthetic datasets (flattening the temporal dimension) and visualize how closely the distributions are in 2D space (Fig.~\ref{fig:performance_aggregation}-c). We observe that overall, the methods also fail to cover the underlying data distribution. DoppelGANger \cite{lin2020using} is the only method that can model metadata, i.e., fault type in this case, and can generate sensible results. However, it still struggles to learn the distribution of the metadata (Fig.~\ref{fig:performance_aggregation}-b). One of the main challenges from this time-series dataset is its size (dimensionality and sequence lengths). Compared to the datasets from their original papers, fast-sampled PMU data from \texttt{PSML} nearly double or triple the number of observations at 960 observations with 91 channels for measurements including voltage magnitude, voltage phase angle, current magnitude, current phase angle, real power, reactive power, and frequency. Current generation approaches leveraging recurrent networks do a poor job of modelling the long-term temporal correlations seen in the data. For long time series, RNNs take many passes to generate the entire samples, which causes them to forget long temporal correlation. Besides, one particular challenge in the power grid data is that each dimension of the time series cannot be handled separately, since the whole system is governed by the  Kirchhoff’s voltage and current laws at each snapshot. This provides an interesting direction for future generation work to address not only the scalability problem but also the constrained generation problem.

\section*{Usage Notes}
The dataset and codes are licensed under the CC BY 4.0, meaning everyone can use it only for non-commercial research purpose.
We recommend users to follow the guidance on Github~\cite{PSML_github_code}, including the details of installation, package usage, dataset navigation, and code navigation. 

\section*{Code availability}
A step-by-step guidance and the source-code for dataset generation and machine learning benchmarks can be found on GitHub~\cite{PSML_github_code}. Specifically, we provide ready-to-use Pytorch data loaders with both data processing and splitting included, and also share the code of evaluators to support fair comparison among different ML-based algorithms, of which the dependencies and usage are also descibed on Github~\cite{PSML_github_code}({\url{https://github.com/tamu-engineering-research/Open-source-power-dataset}}).

\bibliography{main.bbl}



\section*{Author contributions statement}
Conceptualization, L.X. and Y.L.; Data collection, X.Z.; Co-simulation platform, D.W., T.H. and X.Z.; Benchmark, N.X. and L.T.; Writing - original draft, X.Z., S.S., N.X., L.T. and D.W.; Writing - review and editing, all authors; Visualization, X.Z., N.X. and L.T.; Project administration, L.X. and Y.L.

\section*{Competing interests}
The authors declare no competing interests.

\section*{Figures \& Tables}

\newpage

\begin{table}
\floatbox[{\capbeside\thisfloatsetup{capbesideposition={left,top}}}]{table}[\FBwidth]
{\caption{Classification performance on simulation measurements for event detection, classification and localization. We present $avg \pm stdev$ values for experiments with $10$ random seeds. \vspace{0.2cm}}
\label{tab:classification_performance}} 

\begin{adjustbox}{width=0.8\columnwidth}\begin{tabular}{@{}llccc@{}} \toprule \multirow{2}{*}{Categories} & \multirow{2}{*}{Methods} & Classification $\uparrow$ & Localization $\uparrow$&Detection $\downarrow$ \\ && (Balanced Acc)& (Balanced Acc) & (Macro MAE) \\\midrule \multirow{1}{*}{\begin{tabular}[c]{@{}l@{}}Power Domain\end{tabular}} 
& PMU score \cite{pandey2020real} &\textemdash & 0.266 &\textemdash\\
\midrule 
\multirow{3}{*}{\begin{tabular}[c]{@{}l@{}}Traditional \\ Machine \\ Learning \\ Methods\end{tabular}} & 1-NN Euclidean~\cite{bagnall2018uea,dau2019ucr} &0.537 & 0.402&36.465 \\\ & 1-NN DTW-i~\cite{bagnall2018uea,dau2019ucr} & 0.610 & 0.463 &53.928 \\ & 1-NN DTW-d~\cite{bagnall2018uea,dau2019ucr} & 0.598&\textbf{0.474} & 53.709 \\ &MiniRocket~\cite{dempster2021minirocket}&0.690 $\pm$ 0.022&0.208 $\pm$ 0.226&53.908 $\pm$ 3.358\\\midrule \multirow{7}{*}{\begin{tabular}[c]{@{}l@{}}\\Convolutional \\ Neural \\ Networks\end{tabular}} & Vanilla CNN & 0.564 $\pm$ 0.058 & 0.168 $\pm$ 0.053 & 40.458 $\pm$ 12.686 \\ & InceptionTime~\cite{fawaz2020inceptiontime} &0.715 $\pm$ 0.040 & 0.243 $\pm$ 0.047&43.743 $\pm$ 10.605 \\ & MLSTM-FCN~\cite{karim2019multivariate} & \textbf{0.742 $\pm$ 0.029} & 0.285 $\pm$  0.023&\textbf{31.873 $\pm$ 5.400} \\ & ResNet~\cite{wang2017time} & 0.725 $\pm$ 0.049 & 0.232 $\pm$ 0.044 & 38.578 $\pm$ 9.569 \\ & MC-DCNN~\cite{zheng2014time} & 0.726 $\pm$ 0.019 & 0.437 $\pm$ 0.030&38.107 $\pm$ 5.675 \\ & TapNet~\cite{zhang2020tapnet} & 0.653 $\pm$ 0.018 & 0.397 $\pm$ 0.065&58.251 $\pm$ 1.974 \\\midrule \multirow{5}{*}{\begin{tabular}[c]{@{}l@{}}\\Other \\ Deep \\ Learning \\ Methods\end{tabular}} & Fully-connected Neural Network & 0.583 $\pm$ 0.042& 0.245 $\pm$ 0.035 & 54.131 $\pm$ 9.964 \\ & Vanilla RNN & 0.504 $\pm$ 0.045 & 0.224 $\pm$ 0.037 &57.184 $\pm$ 4.285 \\ & LSTM~\cite{hochreiter1997long} & 0.544 $\pm$ 0.049 & 0.248 $\pm$ 0.043&56.434 $\pm$ 2.851 \\ & GRU~\cite{chung2014empirical} & 0.653 $\pm$ 0.029 & 0.332 $\pm$ 0.062&55.550 $\pm$ 2.090 \\ & Vanilla Transformer~\cite{vaswani2017attention} & 0.612 $\pm$ 0.041&0.340 $\pm$ 0.090 &46.824 $\pm$ 0.866 \\ \bottomrule \end{tabular}
\end{adjustbox}
\vspace{-0.2cm}
\end{table}

\begin{table}
\floatbox[{\capbeside\thisfloatsetup{capbesideposition={left,top}}}]{table}[\FBwidth]
{\caption{Performance on 1-hour-ahead load point forecast and interval prediction. \vspace{0.1cm}}
\label{tab:forecasting_performance_load_t+60}}

\begin{adjustbox}{width=\columnwidth}
\begin{tabular}{llccccccccccccc} \toprule \multirow{3}{*}{Categories} & \multirow{3}{*}{Methods} & \multicolumn{9}{c}{Point Forecast} & & \multicolumn{3}{c}{95\% Pred. Interval} \\\cmidrule(l){3-15} & &  \multicolumn{3}{c}{RMSE $\downarrow$} & \multicolumn{3}{c}{MAE $\downarrow$} & \multicolumn{3}{c}{MAPE $\downarrow$} & & \multicolumn{3}{c}{MIS $\downarrow$} \\\cmidrule(l){3-15} & & 2018 & 2019 & 2020 & 2018 & 2019 & 2020 & 2018 & 2019 & 2020 & & 2018 & 2019 & 2020 \\ \midrule \multirow{3}{*}{\begin{tabular}[c]{@{}l@{}}Time-series \\ Models\end{tabular}} & Naive & 0.041&0.039&0.032&0.029&0.029&0.024&0.032&0.032&0.027&&0.203&0.175&0.190 \\ & ARIMA & 0.031&0.027&0.024&0.022&0.020&\textbf{0.017}&0.023&0.022&0.019&&0.113&0.095&0.083\\ & ETS &0.029&\textbf{0.026}&\textbf{0.022}&\textbf{0.021} &\textbf{0.019}&\textbf{0.017}&0.022&\textbf{0.021}&\textbf{0.018}&&0.107&\textbf{0.088}&\textbf{0.076}\\\midrule \multirow{4}{*}{\begin{tabular}[c]{@{}l@{}}Traditional \\ Machine \\ Learning \\ Methods\end{tabular}} & SVR & 0.057&0.058&0.044&0.046&0.048&0.037&0.053&0.057&0.043&&0.181&0.182&0.185\\ & RF &0.033&0.032&0.028&0.026&0.026&0.022&0.029&0.028&0.024&&0.099&0.093&0.095 \\ & GBDT & 0.033&0.032&0.024&0.026&0.026&0.020&0.029&0.029&0.023&&\textbf{0.095}&0.087&0.091 \\ & LR &\textbf{0.026}&0.027&0.030&0.020&0.022&0.024&\textbf{0.021}&0.024&0.026&&1.141&0.871&0.100 \\\midrule \multirow{3}{*}{\begin{tabular}[c]{@{}l@{}}Multilayer \\ Perceptron\end{tabular}} & ELM & 0.229&0.145&0.184&0.196&0.119&0.146&0.233&0.140&0.170&&0.531&0.427&0.573 \\\ & FNN & 0.088&0.091&0.114&0.073&0.076&0.085&0.083&0.089&0.100&&0.240&0.248&0.257 \\ & N-BEATS~\cite{oreshkin2019n} &0.090&0.091&0.084&0.067&0.072&0.068&0.072&0.082&0.077&&0.209&0.207&0.236 \\\midrule \multirow{3}{*}{\begin{tabular}[c]{@{}l@{}}Convolutional \\ Neural \\ Networks\end{tabular}} & Vanilla CNN & 0.103&0.065&0.089&0.085&0.049&0.058&0.099&0.053&0.067&&0.263&0.243&0.243 \\ & WaveNet~\cite{oord2016wavenet} & 0.159&0.109&0.123&0.132&0.087&0.103&0.156&0.093&0.120&&0.438&0.454&0.659 \\ & TCN~\cite{lea2017temporal} &1.336&0.966&1.133&1.330&0.960&1.130&1.478&1.100&1.277&&2.777&1.996&2.359 \\\midrule \multirow{5}{*}{\begin{tabular}[c]{@{}l@{}}Recurrent\\ Neural \\ Networks\end{tabular}} & Vanilla RNN& 0.154&0.094&0.084&0.122&0.075&0.068&0.137&0.088&0.080&&0.330&0.273&0.230 \\\ & LSTM~\cite{hochreiter1997long} & 0.070&0.136&0.167&0.055&0.114&0.101&0.061&0.136&0.120&&0.244&0.286&0.302 \\ & GRU~\cite{chung2014empirical}& 0.087&0.132&0.092&0.071&0.108&0.074&0.081&0.127&0.086&&0.216&0.286&0.201 \\ & LSTNet~\cite{lai2018modeling} & 0.065&0.056&0.0826&0.050&0.046&0.067&0.055&0.053&0.075&&0.188&0.328&0.227 \\ & DeepAR~\cite{salinas2020deepar} &0.068&0.110&0.093&0.055&0.086&0.078&0.059&0.101&0.090&&1.752&1.515&1.531 \\\midrule \multirow{2}{*}{\begin{tabular}[c]{@{}l@{}}Transformer\\ -based \end{tabular}} & Transformer~\cite{vaswani2017attention} & 0.108&0.121&0.132&0.086&0.095&0.101&0.098&0.114&0.119&&0.265&0.280&0.298 \\\ & Informer~\cite{zhou2021informer} & 0.129&0.110&0.073&0.102&0.086&0.059&0.118&0.101&0.067&&0.252&0.215&0.214 \\\midrule - & Neural ODE~\cite{chen2018neural} &0.117&0.265&0.151&0.098&0.249&0.128&0.102&0.270&0.137&&0.737&0.709&0.801 \\ \bottomrule \end{tabular}
\end{adjustbox}

\end{table}

\begin{table}[t]

\floatbox[{\capbeside\thisfloatsetup{capbesideposition={left,top}}}]{table}[\FBwidth]
{\caption{Run-time and fidelity metrics on millisecond-level eventful PMU datasets. Auto- and cross-correlation are calculated as the
sum of the absolute difference between the correlation coefficients computed from real and generated samples. \vspace{0.2cm}}
\label{tab:fidelity}}

\begin{adjustbox}{width=0.65\columnwidth}
\begin{tabular}{@{}lcccc@{}}
\toprule
\multirow{2}{*}{Method} & \multirow{2}{*}{Autocorrelation $\downarrow$} & Cross-                             & Discriminative    & \multirow{2}{*}{Hours $\downarrow$} \\
             &        & correlation $\downarrow$ & Score $\downarrow$                       &  \\ \midrule
NaiveGAN \cite{goodfellow2014generative}     & 663.13 & 7175.78                    & 0.492 $\pm$ 0.022                           & 5\\
RCGAN \cite{esteban2017real}       & 667.89 & 7607.29                    & 0.499 $\pm$ 0.002                           &  27\\
COT-GAN \cite{xu2020cot}     & 112.42 & 2532.69                    & \textbf{0.435 $\pm$ 0.016} & 6  \\
TimeGAN \cite{yoon2019time}                 & \textbf{72.56}                 & \textbf{1361.45} & 0.481 $\pm$ 0.008 & 52                                      \\
DoppelGANger \cite{lin2020using}& 86.76  & 3994.45                    & 0.447 $\pm$ 0.004                           & 22 \\ \bottomrule
\end{tabular}
\end{adjustbox}
\vspace{-0.3cm}
\end{table}

\end{document}